\documentclass[letterpaper, 10 pt, journal, twoside]{IEEEtran}

\IEEEoverridecommandlockouts                              

\makeatletter
\let\NAT@parse\undefined
\makeatother
\usepackage[numbers]{natbib}
\usepackage{etoolbox}

\let\oldthebibliography\thebibliography

\renewcommand{\thebibliography}[1]{
  \oldthebibliography{#1}
  \setlength{\parskip}{0pt}
  \setlength{\itemsep}{-2pt}
  \setlength{\bibsep}{5pt}
}

\usepackage{graphics} 
\usepackage{mathptmx} 
\usepackage{times} 
\usepackage{amsmath} 
\usepackage{amssymb}  
\usepackage{bm}

\usepackage{graphicx}
\usepackage{booktabs}
\usepackage{threeparttable}
\usepackage{multirow}

\usepackage{balance}
\usepackage{rpm_SIunits}
\usepackage{rpm_acronyms}
\usepackage{rpm_math}
\usepackage{rpm_misc}

\usepackage{soul,color}

\usepackage{lipsum}

\usepackage[font=small,labelfont=bf]{caption}
\usepackage{algorithm}
\usepackage{algorithmic}

\usepackage{xcolor}

\usepackage{tikz} 

\usepackage{caption}
\usepackage{subcaption}

\usepackage{hyperref}
\hypersetup{
    colorlinks=true,
    linkcolor=blue,
    filecolor=magenta,
    urlcolor=blue,
    citecolor=black
}

\usepackage{multicol}
\usepackage{float} 
\usepackage{placeins}
\usepackage{url}
\title{\LARGE \bf 

Geometrically-Constrained Radar-Inertial Odometry

via Continuous Point-Pose Uncertainty Modeling

}

\author{
Wooseong Yang$^{1}$, Dongjae Lee$^{1}$, Minwoo Jung$^{1}$, and Ayoung Kim$^{1*}$

\thanks{Manuscript received: November 17, 2025; Revised: January 21, 2026; Accepted: February 28, 2026. This paper was recommended for publication by
Editor Javier Civera upon evaluation of the Associate Editor and Reviewers
comments.}
\thanks{$^\dagger$This work was supported by the Technology Innovation Program (1415187329, 20024355, Development of autonomous driving connectivity technology based on sensor-infrastructure cooperation) funded by the Ministry of Trade, Industry \& Energy (MOTIE, Korea)}%
\thanks{$^{1}$ W. Yang, D. Lee, M. Jung and A. Kim are with the Dept. of Mechanical Engineering, SNU, Seoul, S. Korea {\tt\footnotesize [yellowish, pur22, moonshot, ayoungk]@snu.ac.kr}}%
\thanks{Digital Object Identifier (DOI): see top of this page.}

}

\begin{document}

\maketitle

\begin{abstract} 

Radar odometry is crucial for robust localization in challenging environments; however, the sparsity of reliable returns and distinctive noise characteristics impede its performance.
This paper introduces geometrically-constrained radar-inertial odometry and mapping that jointly consolidates point and pose uncertainty.
We employ the continuous trajectory model to estimate the pose uncertainty at any arbitrary timestamp by propagating uncertainties of the control points. These pose uncertainties are continuously integrated with heteroscedastic measurement uncertainty during point projection, thereby enabling dynamic evaluation of observation confidence and adaptive down-weighting of uninformative radar points.
By leveraging quantified uncertainties in radar mapping, we construct a high-fidelity map that improves odometry accuracy under imprecise radar measurements.
Moreover, we reveal the effectiveness of explicit geometrical constraints in radar-inertial odometry when incorporated with the proposed uncertainty-aware mapping framework.
Extensive experiments on diverse real-world datasets demonstrate the superiority of our method, yielding substantial performance improvements in both accuracy and efficiency compared to existing baselines.
We release our open-source implementation at: \href{https://github.com/wooseongY/GeoRIO}{https://github.com/wooseongY/GeoRIO}.

\end{abstract}


\captionsetup[subfigure]{labelformat=simple}
\vspace{-4.0mm}
\section{Introduction}
\label{sec:intro}

Radar odometry has gained notable attention in robotics due to its robustness in adverse weather conditions \cite{harlow2024anew}, yet it still suffers from data sparsity. Recent approaches address these limitations by scan-to-submap registration, enforcing geometric consistency through denser structural priors from accumulated scans \cite{lee2024lidar}.
While these can also mitigate abrupt drift \cite{vizzo2023kiss}, the distinctive noise characteristics of radar impede reliable submap construction and further lead to erroneous pose estimation. Signal penetrability due to long wavelength exhibits inherent difficulty in distinguishing meaningful measurements from spurious noise \cite{yang2025ground}.

Overcoming this ambiguity, most existing radar odometry and mapping strategies rely primarily on dynamic object filtering using velocity measurements \cite{zhuang20234d, herraez2024radar}. These fail to suppress sensor-induced artifacts, thus contaminating the submap and further degrading scan-to-submap registration \cite{herraez2025rai}. This limitation motivates a rigorous point assessment, for which uncertainty modeling provides a promising approach to quantify the reliability of each measurement \cite{jung2023asynchronous}.

\begin{figure}[t]
    \centering
    \includegraphics[width=0.9\linewidth, trim=0 0.3cm 0.5cm 0.2cm, clip]{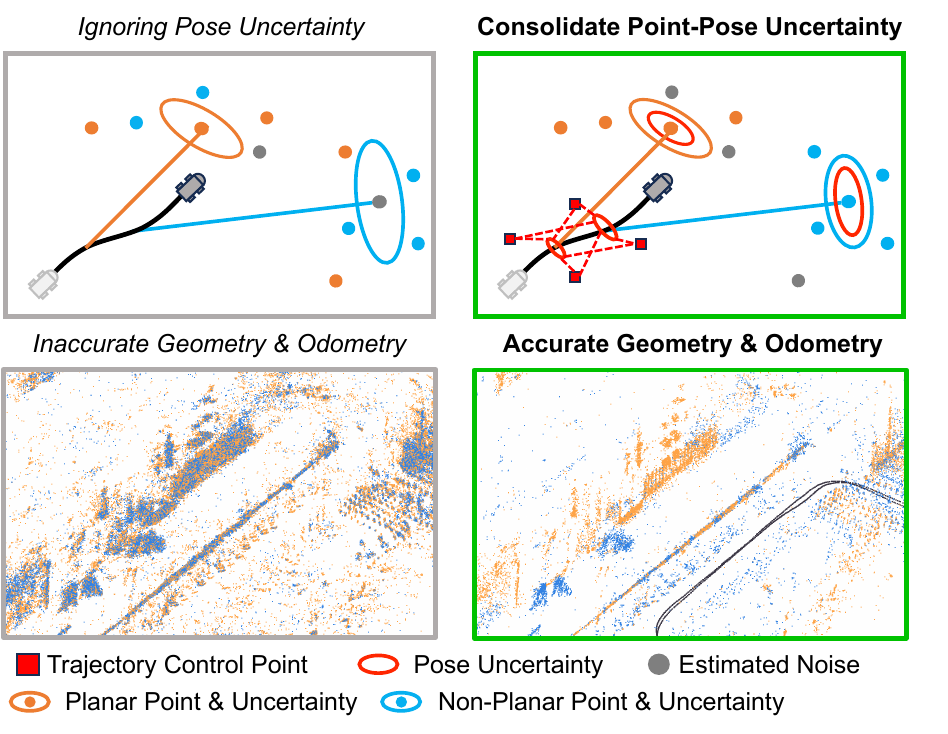}
    \caption{ 
        We jointly model and continuously consolidate point-pose uncertainty in radar odometry and mapping, enabling the effective usage of explicit local geometry.
    }
    \label{fig:teaser}
\vspace{-7mm}
\end{figure}

Two key considerations are indispensable for reliable uncertainty analysis for radar odometry and mapping: (i) modeling heteroscedastic, view-dependent measurement uncertainty \cite{zhang20234d}, (ii) addressing pose-uncertainty-aware propagation, which affects accurate registration and consistent mapping \cite{yuan2022efficient}.
However, prior studies have discussed modeling measurement uncertainty only and have underexplored its propagation during accumulation \cite{xu2025incorporating}.
This omission compromises mapping consistency by failing to account for erroneous point projections stemming from pose drift. Radar's low angular resolution exacerbates this issue, amplifying minor pose estimate errors into significant spatial ambiguities.


To address these limitations, we close this gap by introducing a radar-inertial odometry and mapping framework that jointly addresses both challenges in uncertainty analysis of the radar.
By adopting a continuous trajectory, our method enables pose uncertainty modeling at arbitrary timestamps based on the uncertainty of each control point.
As shown in \figref{fig:teaser}, we then compound the pose uncertainty with each point-wise uncertainty to perform a continuous uncertainty propagation during point projection. Through our uncertainty modeling, we reveal that the use of explicit geometrical constraints, such as planar observation models or localizability analysis, becomes more effective even with sparse and imprecise radar returns. We also achieved a high-fidelity map by selectively retaining points with low uncertainty during map maintenance. Extensive experiments on real-world benchmarks demonstrate that our proposed framework achieves accurate odometry and high-fidelity maps in real-time, outperforming state-of-the-art baselines.
Our contributions are summarized as follows:

\begin{figure*}[!t]
    \centering
    \includegraphics[width=0.85\linewidth, trim=0 0 0 0, clip]{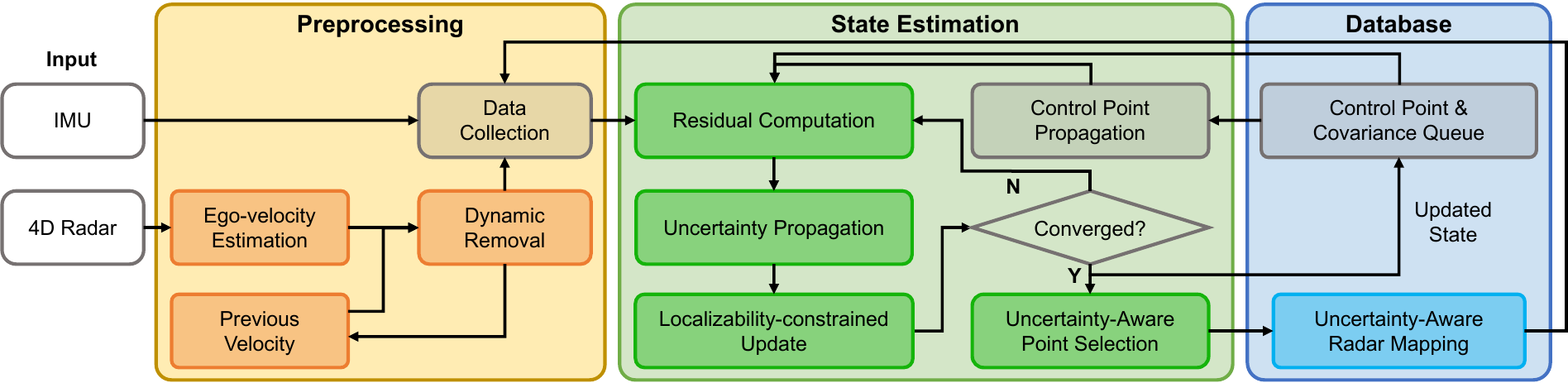}
    \caption{ 
        Our method consists of preprocessing, state estimation, and mapping. In preprocessing, we remove dynamic points from radar measurements using ego-velocity estimates and perform data association for the measurements used in the state update. The state estimation module iteratively performs residual computation, uncertainty propagation, and localizability-constrained IEKF updates until convergence. After convergence, we utilize high-confidence points to enhance mapping accuracy. 
    }
    \vspace{-6mm}
    \label{fig:pipeline}
\end{figure*}


\begin{itemize}    


    \item \textbf{Jointly consolidating point-pose uncertainty in radar odometry and mapping:} We propose a unified framework that jointly propagates the measurement and pose uncertainty in radar odometry and mapping. Our uncertainty analysis enables robust down-weighting of uninformative points in the observation model. Furthermore, we iteratively refine the map by replacing high-uncertainty points with lower-uncertainty measurements, thereby enhancing odometry accuracy while preserving map fidelity.
    
    \item \textbf{Revealing the effectiveness of explicit geometrical constraint in radar odometry:} We demonstrate that the use of explicit local geometric features and localizability analysis, which were overlooked in existing radar odometry baselines, can yield a noticeable performance improvement with our uncertainty-aware approach.
    
    \item \textbf{Extensive experiments in real-world datasets:} We validate our system on diverse datasets, including indoor and outdoor sequences across structured, unstructured, and degenerate environments, demonstrating substantial improvements in accuracy during real-time operation.
\end{itemize}

\section{Related Work}
\label{sec:relatedworks}

\subsection{Radar Odometry with Scan Registration}

With advances in the accuracy of radar returns with automotive radar development, subsequent studies began to exploit point cloud registration in radar odometry. \citet{zhang20234d} proposed an adaptive probabilistic distribution Generalized ICP (APD-GICP) that models measurement covariance during sequential registration. Building on this, \citet{yang2025ground} introduced cluster-based weighting and \citet{xu2025incorporating} modeled point-wise uncertainty to enable probability-guided correspondence, improving registration accuracy. Nevertheless, these methods are based on scan-to-scan matching, which is susceptible to sudden drift in sequential odometry and remains vulnerable to noise-dominant measurements due to sparsity and limited \ac{FOV}.

To mitigate these issues, several methods integrate local maps into radar odometry. \citet{li20234d} performed \ac{NDT}-based scan-to-submap registration, while \citet{zhuang20234d} adopted point-to-distribution residuals to improve alignment. \citet{wu2024efear} proposed correspondence weighting strategies to reduce the impact of outliers during submap matching. \citet{herraez2024radar, herraez2025rai} refined scan-to-submap matching by incorporating Doppler velocity and further leveraging ground information to enhance localization precision \cite{herraez2025ground}. However, many of these approaches accumulated scans without quantifying the measurement reliability during map construction. As a result, artifacts often persisted in the aggregated map, degrading odometry accuracy. We therefore propose a reliability quantification methodology with novel uncertainty modeling over the aggregated map, thereby improving map fidelity and odometry performance.

\subsection{Uncertainty in Range-Sensor SLAM} 

Uncertainty in \ac{SLAM} has been classically formalized as the covariance of the estimated state \cite{kaess2008smoothing}. Recently, sophisticated uncertainty analysis has become central to \ac{LiDAR} \ac{SLAM}. For instance, \citet{yuan2022efficient} propagated uncertainty into adaptive voxel mapping, while \citet{jung2023asynchronous} incorporated uncertainty to mitigate inter-sensor discrepancies. \citet{huang2024loglio2} proposed uncertainty analysis by exploiting surface characteristics. These advancements address uncertainties by quantifying confidence in poses and landmarks, thereby enhancing the robustness of point matching and odometry.

In contrast, the treatment of uncertainty is absent in the radar \ac{SLAM} domain. While \citet{xu2025incorporating} incorporated per-point uncertainty into radar registration, their approach omitted uncertainty propagation to the global map level, which could lead to biased or noisy landmarks. To address this issue, we propose the radar-inertial odometry and mapping with compounded uncertainty analysis.

\section{Preliminary}
\label{sec:method}

The lower letter without bold represents the scalar (e.g., a), the lower letter with bold represents the vector (e.g., \textbf{a}), and the upper letter with bold represents the matrix (e.g., \textbf{A}).

We adopt the \ac{IEKF}-based continuous trajectory estimation model with cubic B-spline \cite{cao2025resple}. $\mathbf{t}(t)\in\mathbb{R}^3$ and $\mathbf{q}(t)\in\mathbb{S}^3$ denote the translation and orientation in quaternion at time $t\in\left[t_{i-1},t_i\right)$. These state variables are parametrized by control points $\underline{\mathbf{t}}_n\in\mathbb{R}^3, \underline{\mathbf{q}}_n\in\mathbb{S}^3$ at knots (same as timestamp) $t_n \left(n=i-3,\cdots,i\right)$ in fixed time interval $\Delta t$ as follows:

\vspace{-2mm}\small
\begin{equation}
\label{eq:translation}
    \mathbf{t}(t)=\left[\underline{\mathbf{t}}_{i-3},\underline{\mathbf{t}}_{i-2},\underline{\mathbf{t}}_{i-1},\underline{\mathbf{t}}_i\right]\mathbf{M}\underline{\mathbf{u}}, \quad \underline{\mathbf{u}} = [1,u,u^2,u^3]^{\top},
\end{equation}
\normalsize
where $\mathbf{M} = \frac{1}{6}
                        \left[\begin{smallmatrix}
                        1 & -3 & 3 & -1 \\
                        4 & 0 & -6 & 3 \\
                        1 & 3 & 3 & -3 \\
                        0 & 0 & 0 & 1
                        \end{smallmatrix}\right]$ is the basis matrix and $u=(t-t_{i-1})/\Delta t$.
The quaternion-based orientation is represented as the following cumulative equation:

\vspace{-2mm}\small
\begin{equation}
\label{eq:orientation}
    \mathbf{q}(t) = \underline{\mathbf{q}}_{i-4}\otimes\prod_{k=i-3}^i \mathrm{Exp}(\lambda_k\underline{\boldsymbol{\delta}}_k),
\end{equation}
\normalsize
where $\otimes$ denotes the quaternion multiplication, and $\underline{\boldsymbol{\delta}}_k=\mathrm{Log}(\underline{\mathbf{q}}_{k-1}^{-1}\otimes\underline{\mathbf{q}}_k)$ is the increment of the orientation control points. The cumulative basis function $\lambda_k$ are given {as} follows:

\vspace{-2mm}\small
\begin{equation}
    [\underline{\lambda}_{i-3},\underline{\lambda}_{i-2},\underline{\lambda}_{i-1},\underline{\lambda}_i]^\top = \widetilde{\mathbf{M}}\underline{\mathbf{u}},
\end{equation}
\normalsize
where $\widetilde{\mathbf{M}} = \frac{1}{6}\left[\begin{smallmatrix}
        6 & 0 & 0 & 0 \\
        5 & 3 & -3 & 1 \\
        1 & 3 & 3 & -2 \\
        0 & 0 & 0 & 1
    \end{smallmatrix}\right]$ is the cumulative blending matrix.

Our radar-inertial system utilizes the spline-state-space modeling with \ac{IMU} biases as below:

\vspace{-4mm}\small
\begin{align}
\label{eq:state}
    \mathbf{x}_k&= [{\mathbf{x}_k^p}^\top, {\mathbf{x}_k^q}^\top,\mathbf{b}_a^\top,\mathbf{b}_g^\top]^\top \in\mathbb{R}^{30}, \\
    \mathbf{x}_k^p&=\left[\underline{\mathbf{t}}_{i-3}^\top,\underline{\mathbf{t}}_{i-2}^\top,\underline{\mathbf{t}}_{i-1}^\top,\underline{\mathbf{t}}_i^\top\right]^\top\in\mathbb{R}^{12}, \\ 
    \mathbf{x}_k^q&=\left[\underline{\boldsymbol{\delta}}_{i-3}^\top,\underline{\boldsymbol{\delta}}_{i-2}^\top,\underline{\boldsymbol{\delta}}_{i-1}^\top,\underline{\boldsymbol{\delta}}_i^\top\right]^\top\in\mathbb{R}^{12},
\end{align}
\normalsize
where $\textbf{x}^p$ and $\textbf{x}^q$ indicate the translation control points and the increment of orientation control points, and $\mathbf{b}_a$ and $\mathbf{b}_g$ are acceleration and gyroscope biases, respectively.

The prediction step of IEKF is performed as follows:

\vspace{-4mm}\small
\begin{align}
    \mathbf{x}_{k|k-1} &= \mathbf{A}_{k-1}\mathbf{x}_{k-1|k-1}, \\
    \mathbf{P}_{k|k-1} &= \mathbf{A}_{k-1}\mathbf{P}_{k-1|k-1}\mathbf{A}_{k-1}^\top+\mathbf{Q}_{k-1},
\end{align}
\normalsize
with the state mean $\mathbf{x}_{k-1|k-1}$ and the corresponding covariance $\mathbf{P}_{k-1|k-1}$ from the previous step. $\mathbf{Q}_{k-1}$ denotes the propagation noise term.
We adopt transition matrix following \cite{cao2025resple} with translation and rotation submatrices $\mathbf{A}^p$ and $\mathbf{A}^q$:

\vspace{-4mm}\small
\begin{align}
    \mathbf{A}_{k-1} &= diag\left[\mathbf{A}^p,\mathbf{A}^q,\mathbf{I},\mathbf{I}\right], \\
    \mathbf{A}^p &= \begin{bmatrix}
                        \mathbf{0} & \mathbf{I} & \mathbf{0} & \mathbf{0} \\
                        \mathbf{0} & \mathbf{0} & \mathbf{I} & \mathbf{0} \\
                        \mathbf{0} & \mathbf{0} & \mathbf{0} & \mathbf{I} \\
                        -\mathbf{I} & \mathbf{0} & 2\mathbf{I} & \mathbf{0}
                        \end{bmatrix},
    \mathbf{A}^q = \begin{bmatrix}
                        \mathbf{0} & \mathbf{I} & \mathbf{0} & \mathbf{0} \\
                        \mathbf{0} & \mathbf{0} & \mathbf{I} & \mathbf{0} \\
                        \mathbf{0} & \mathbf{0} & \mathbf{0} & \mathbf{I} \\
                        \mathbf{0} & \mathbf{I} & \mathbf{0} & \mathbf{0}
                        \end{bmatrix}.
\end{align}
\normalsize

Then the Kalman gain $\mathbf{K}_j$ and motion increment are computed as follows:

\vspace{-4mm}\small
\begin{align}
    \mathbf{K}_j &=(\mathbf{H}_j\mathbf{R}_j\mathbf{H}_j^{-1}+\mathbf{P}_{k|k-1}^{-1})^{-1}\mathbf{H}_j^\top\mathbf{R}_j^{-1},\\
    \delta\mathbf{x}_{j|j} &= \mathbf{K}_j(z_j-h(\mathbf{x}_j))-(\mathbf{I}-\mathbf{K}_j\mathbf{H}_j)(\mathbf{x}_j-\mathbf{x}_{k|k-1}).
\end{align}\normalsize
\noindent with the observation model $z_j-h(\mathbf{x}_j)=0$, Jacobian $\mathbf{H}_j$, and measurement covariance $\mathbf{R}_j$.

The iterative update step is initialized with $\mathbf{x}_j=\mathbf{x}_{k|k-1}$ at {\hbox{$j=0$}}. At the $j$-th step, the state is updated by $\mathbf{x}_{j+1}=\mathbf{x}_j + \delta\mathbf{x}_{j+1|j}$, while $\delta\mathbf{x}_{j+1|j}$ denotes the projection of state increment, which will be described in \secref{secref:localizability}.
At every update step, we utilize the radar and \ac{IMU} measurements between the last updated control point and the newly added control point.

\section{Methodology}
This section presents the details of our algorithm, which is illustrated in \figref{fig:pipeline}. We first preprocess each radar scan with ego-velocity-based dynamic removal \cite{jung2024coral}, and stabilize the dynamic filtering by reusing the previous ego-velocity when abrupt changes occur \cite{herraez2024radar}. The remaining subsections describe our localizability-constrained update, continuous-time point-pose uncertainty propagation, scan-to-submap registration using explicit local geometry, and \ac{IMU} residuals.

\subsection{Localizability-Constrained IEKF update}
\label{secref:localizability}

Unlike existing methods, we construct a localizability Hessian and regularize the control point update using a constrained IEKF to handle geometrical degeneracy.
For the point $\mathbf{p}\in\mathbb{R}^3$ with its normal vector $\mathbf{n}\in\mathbb{R}^3$, we {derive} the Hessian matrix {from \cite{pomerleau2024areview}} for localizability estimation:

\vspace{-2mm}\small
\begin{equation}
    \boldsymbol{\Lambda}=
    \begin{bmatrix}
        \boldsymbol{\Lambda}_{tt} & \boldsymbol{\Lambda}_{rt} \\
        \boldsymbol{\Lambda}_{tr} & \boldsymbol{\Lambda}_{rr}
    \end{bmatrix}
    =\mathbf{L}^\top\mathbf{L}\in\mathbb{R}^{6\times6}, \mathbf{L} = [\mathbf{n}^\top,(\mathbf{p}\times\mathbf{n})^\top],
\end{equation}
\normalsize
where the $\boldsymbol{\Lambda}_{tt},\boldsymbol{\Lambda}_{rr}\in\mathbb{R}^{3\times3}$ denote the translational and rotational blocks of the localizability Hessian, respectively. With their eigen matrix $\mathbf{E}_t,\mathbf{E}_r\in\mathbb{R}^{3\times3}$, we define its 6-DoF localizability score vector $\mathbf{s}(\mathbf{p})=\left[\mathbf{n}^\top\mathbf{E}_t, \left(\frac{\mathbf{p}\times\mathbf{n}}{\left\Vert\mathbf{p}\times\mathbf{n}\right\Vert_2}\right)^\top\mathbf{E}_r\right]$. The point $\mathbf{p}$ is deemed informative on $i$-axis if $|s_i|>\eta$. If the number of informative points for an axis is insufficient, we regard that axis as under-constrained.
Let $\mathbf{C}$ be the constraint matrix whose rows are the eigenvectors corresponding to the under-constrained axes, so that the admissible update satisfies $\mathbf{C}\,\delta\mathbf{x}=\mathbf{0}$.
To suppress update components along these degenerate directions, we project the motion increment onto {the null space $\mathcal{N}(\mathbf{C})$ of the constraint matrix $\mathbf{C}$} as follows:

\vspace{-4mm}\small
\begin{align}
\delta\mathbf{x}_{j+1\mid j}[18:]=\left(\mathbf{I}-\boldsymbol{\Upsilon}\mathbf{C}\right)\delta\mathbf{x}_{j\mid j}[18:],\quad
\boldsymbol{\Upsilon}=\mathbf{C}^{\top}(\mathbf{C}\mathbf{C}^{\top})^{-1}.
\end{align}
\normalsize
We iterate until $||\delta\mathbf{x}_{j+1|j}||_2<\epsilon$, then update the state and covariance as:

\vspace{-2mm}\small
\begin{equation}
    \mathbf{x}_{k|k}=\mathbf{x}_{j+1}, \mathbf{P}_{k|k}=(\mathbf{I}-\mathbf{K}_j\mathbf{H}_j)\mathbf{P}_{k|k-1}.
\end{equation}
\normalsize

\subsection{Continuous Propagation of Point-Pose Uncertainties}



{Compared to the discrete-time formulation, the continuous-time trajectory effectively accommodates asynchronous radar and \ac{IMU} measurements by enabling pose estimation and world-frame transformation of each radar return at its measurement timestamp. However, adopting continuous-time representation alone is not guaranteed to improve odometry accuracy, as shown in \tabref{tab:ablation_unc}. This is because the reliability of each transformed point is jointly governed by (i) pose uncertainty and (ii) sensor measurement uncertainty.} In this section, we derive these uncertainty terms from our state models and describe their continuous propagation processes.

According to \eqref{eq:translation}, $\mathbf{t}(t)$ can be rewritten as follows:

\vspace{-4mm}\small
\begin{align}
    \mathbf{t}(t)&=\mathbf{B}(u)\begin{bmatrix}
        \underline{\mathbf{t}}_{i-3}^\top, 
        \underline{\mathbf{t}}_{i-2}^\top,
        \underline{\mathbf{t}}_{i-1}^\top,
        \underline{\mathbf{t}}_{i}^\top
    \end{bmatrix}^\top, \\
    \mathbf{B}(u) &= [m_0(u)\mathbf{I},m_1(u)\mathbf{I},m_2(u)\mathbf{I},m_3(u)\mathbf{I}]\in \mathbb{R}^{3\times12}, \\
    \mathbf{M}\underline{\mathbf{u}}&=[m_0(u),m_1(u),m_2(u),m_3(u)]^\top,
\end{align}
\normalsize
while $\mathbf{B}(u)$ is the Jacobian of translation w.r.t. control points. Let the uncertainty of translation control point be $\boldsymbol{\Sigma}_{\underline{\mathbf{t}}_k}$. Then, we can estimate the uncertainty of the translation as follows:

\vspace{-4mm}\small
\begin{align}
\label{eq:p_unc}
    \boldsymbol{\Sigma}_{\mathbf{t}(t)}&=\mathbf{B}(u)\begin{bmatrix}
        \boldsymbol{\Sigma}_{\underline{\mathbf{t}}_{i-3}} & \cdots & \mathbf{0} \\
        \vdots & \ddots & \vdots \\
        \mathbf{0} & \cdots & \boldsymbol{\Sigma}_{\underline{\mathbf{t}}_i}
    \end{bmatrix} \mathbf{B}(u)^\top \nonumber \\
    &=\sum_{k=0}^3(m_k(u))^2\boldsymbol{\Sigma}_{\underline{\mathbf{t}}_{i-3+k}}.
\end{align}
\normalsize

Similarly, let the uncertainty of the orientation control point $\underline{\boldsymbol{\delta}}_i$ be $\boldsymbol{\Sigma}_{\underline{\boldsymbol{\delta}}_i}$. According to \eqref{eq:orientation}, orientation $\mathbf{q}$ and its Jacobian w.r.t. $\underline{\boldsymbol{\delta}}_j$ can be rewritten as follows:

\vspace{-4mm}\small
\begin{align}
    \mathbf{q}(t) &= \underline{\mathbf{q}}_{i-4}\otimes \prod_{k=i-3}^{j-1} \mathrm{Exp}(\lambda_k\underline{\boldsymbol{\delta}}_k) \otimes \mathrm{Exp}(\lambda_j\underline{\boldsymbol{\delta}}_j) \otimes \prod_{k=j+1}^i \mathrm{Exp}(\lambda_k\underline{\boldsymbol{\delta}}_k) \nonumber \\
    &= \left[\underline{\mathbf{q}}_{i-4}\otimes \prod_{k=i-3}^{j-1} \mathrm{Exp}(\boldsymbol{\nu}_k)\right]_L
    \left[\prod_{k=j+1}^i \mathrm{Exp}(\boldsymbol{\nu}_k)\right]_R \mathrm{Exp}(\boldsymbol{\nu}_j), \nonumber \\ 
\label{eq:Jacobian_q_cp}
    \mathbf{J}_{\mathbf{q},j} &= \cfrac{\partial\mathbf{q}}{\partial\underline{\boldsymbol\delta}_j} \nonumber \\
    &= \lambda_j\left[\underline{\mathbf{q}}_{i-4}\otimes \prod_{k=i-3}^{j-1} \mathrm{Exp}(\boldsymbol{\nu}_k)\right]_L
    \left[\prod_{k=j+1}^i \mathrm{Exp}(\boldsymbol{\nu}_k)\right]_R\cfrac{\partial\mathrm{Exp}(\boldsymbol{\nu}_j)}{\partial\boldsymbol{\nu}_j},
\end{align}
\normalsize
where $\boldsymbol{\nu}_l=\lambda_l\underline{\boldsymbol{\delta}}_l$ and $\left[ \cdot \right]_L$ and $\left[ \cdot \right]_R$ denote the left and right matrix representations of the quaternion multiplication.
Moreover, because our quaternion parameterization accumulates over successive states, the orientation of the lagged state $\underline{\mathbf{q}}_{i-4}$ must be addressed. The Jacobian w.r.t. the lagged quaternion can be computed as follows:

\vspace{-4mm}\small
\begin{align}
\label{eq:Jacobian_q_idle}
    \mathbf{q}(t) &= \left[\prod_{k=i-3}^i \mathrm{Exp}(\boldsymbol{\nu}_k)\right]_L \underline{\mathbf{q}}_{i-4} ,\nonumber \\
    \mathbf{J}_{\mathbf{q}, (i-4)} &= \cfrac{\partial\mathbf{q}}{\partial\underline{\mathbf{q}}_{i-4}} = \left[\prod_{k=i-3}^i \mathrm{Exp}(\boldsymbol{\nu}_k)\right]_L.
\end{align}
\normalsize
With \eqref{eq:Jacobian_q_cp} and \eqref{eq:Jacobian_q_idle}, the uncertainty of the estimated orientation can be computed as follows:

\vspace{-2mm}\small
\begin{equation}
\label{eq:rot_unc}
    \boldsymbol{\Sigma}_{\mathbf{q}(t)}=\sum_{k=i-3}^i\left(\mathbf{J}_{\mathbf{q},k}\boldsymbol{\Sigma}_{\underline{\boldsymbol{\delta}}_i}\mathbf{J}_{\mathbf{q},k}^\top\right) + \mathbf{J}_{\mathbf{q},(i-4)} \boldsymbol{\Sigma}_{\underline{\mathbf{q}}_{i-4}}\mathbf{J}_{\mathbf{q},(i-4)}^\top.
\end{equation}
\normalsize

\begin{figure}[t]
    \centering
    \includegraphics[width=0.9\columnwidth, trim=0.05cm 0.125cm 0cm 0, clip]{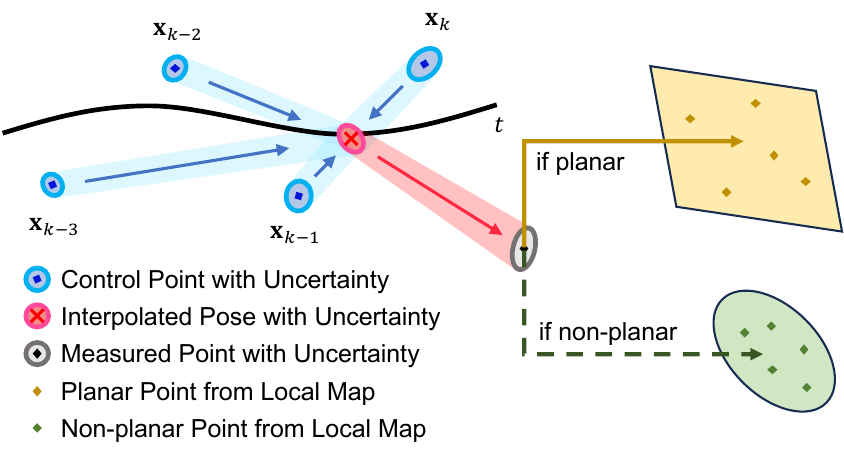}
    \caption{ 
        Conceptual diagram of our uncertainty propagation module. Uncertainties of B-spline control points are propagated to the pose at the radar timestamp and subsequently to individual radar returns. The resulting per-point uncertainties in the global frame perform as confidence weights in the observation model for local geometry-guided scan-to-submap registration.
    }
    \label{fig:propagation}
    \vspace{-6mm}
\end{figure}

Now, assume that we have the measured point expressed in the radar frame ${}^\mathrm{R}\mathbf{p}$ at time $t$, which has measurement uncertainties $\sigma_r$, $ \sigma_a$, and $\sigma_e$ for the range, azimuth, and elevation, respectively. These are propagated to Cartesian coordinates, with the following equation:

\small
\begin{equation}
\label{eq:point_unc}
    \boldsymbol{\Sigma}_{}^\mathrm{R}\mathbf{p}= \boldsymbol{\Gamma}\begin{bmatrix}
        \sigma_r^2 & 0 & 0 \\
        0 & \sigma_a^2 & 0 \\
        0 & 0 & \sigma_e^2
    \end{bmatrix} \boldsymbol{\Gamma}^\top,
\end{equation}
\normalsize
where $\boldsymbol{\Gamma}$ is the Jacobian of the spherical-coordinate mapping.
Then the measured point can be projected to the world frame:

\vspace{-4mm}\small
\begin{align}
    {}^\mathrm{W}\mathbf{p} &= \mathbf{t}(t)+\mathbf{R}(t) \left(\mathbf{R}_\mathrm{IR}{}^\mathrm{R}\mathbf{p}+{}^\mathrm{I}\textbf{t}_\mathrm{IR}\right) \nonumber \\
    &= \mathbf{t}(t)+\mathbf{R}(t){}^\mathrm{I}\mathbf{p},
\end{align}
\normalsize
with extrinsic parameter between the body frame and radar frame $\mathbf{R}_\mathrm{IR},{}^\mathrm{I}\textbf{t}_\mathrm{IR}$. $\mathbf{R}(t)$ denotes the orientation in SO(3) manifold.
Finally, the uncertainty of the point measurement in the world frame can be achieved from \eqref{eq:p_unc}, \eqref{eq:rot_unc}, and \eqref{eq:point_unc}:

\vspace{-4mm}\small
\begin{align}
    \boldsymbol{\Sigma}_{{}^\mathrm{I}\mathbf{p}} &= \mathbf{R}_\mathrm{IR}\boldsymbol{\Sigma}_{{}^\mathrm{R}\mathbf{p}}\mathbf{R}_\mathrm{IR}^\top, \\
\label{eq:final_unc}
    \boldsymbol{\Sigma}_{{}^\mathrm{W}\mathbf{p}} &= \boldsymbol{\Sigma}_{\mathbf{t}(t)}+\mathbf{R}(t)[{}^\mathrm{I}\mathbf{p}]_\times\boldsymbol{\Sigma}_{\mathbf{R}(t)}[{}^\mathrm{I}\mathbf{p}]_\times^\top\mathbf{R}(t)^\top + \mathbf{R}(t)\boldsymbol{\Sigma}_{{}^\mathrm{I}\mathbf{p}}\mathbf{R}(t)^\top.
\end{align}
\normalsize
The uncertainty in SO(3) $\left(\boldsymbol{\Sigma}_{\mathbf{R}(t)}\right)$ can be computed from the uncertainty in quaternion space \eqref{eq:rot_unc}.
With \eqref{eq:final_unc}, we can consolidate both uncertainties that affect the point projection and propagation into the global map frame. \figref{fig:propagation} illustrates the conceptual diagram of our uncertainty propagation method.

 \begin{table*}[t]
\centering
\caption{Evaluation Results in \texttt{HeRCULES}}
\label{tab:result_hercules}
\resizebox{0.9\textwidth}{!}{
\begin{tabular}{c|ccc|ccc|ccc|ccc|ccc}
\toprule
\multicolumn{1}{l|}{\textbf{}} & \multicolumn{3}{c|}{\texttt{Parking Lot 03}} & \multicolumn{3}{c|}{\texttt{Library 01}} & \multicolumn{3}{c|}{\texttt{Sports Complex 01}} & \multicolumn{3}{c|}{\texttt{Street 01}} & \multicolumn{3}{c}{\texttt{Bridge 01}} \\

\multicolumn{1}{l|}{}          & ATE            & RPE$_t$               & RPE$_r$         & ATE            & RPE$_t$         & RPE$_r$         & ATE                  & RPE$_t$         & RPE$_r$         & ATE             & RPE$_t$         & RPE$_r$         & ATE            & RPE$_t$         & RPE$_r$          \\ \midrule
Fast-LIO2 (LiDAR)  & {2.345} & 0.287 & 0.432 & {10.372} & {0.142} & 0.480 & {10.884} & {0.191} & 1.004 & {3.784} & {0.298} & 0.273 & 412.212 & 1.730 & {0.312} \\ \midrule
Go-RIO        & 5.744 & 0.459 & \underline{0.221} & 22.514 & 1.083 & 0.509 & 29.275 & 0.233 & 1.190 & -      & -      & -      & \underline{248.884} & \underline{0.839} & \underline{1.173} \\
PUA-RIO       & 6.111 & 0.390 & 0.937           & 296.136 & 1.876 & 0.856 & \underline{20.816} & 0.269 & 1.280 & 39.024 & 0.371 & 1.184 & -      & -      & -      \\
EKFRIO-TC     & 4.503 & \textbf{0.160} & 0.331 & \underline{17.491} & \underline{0.239} & \underline{0.255} & 20.848 & \underline{0.218} & \underline{0.964} & \underline{24.662} & \underline{0.318} & \underline{0.272} & -      & -      & -      \\

Ours          & \textbf{1.054} & \underline{0.208} & \textbf{0.127} & \textbf{1.555} & \textbf{0.090} & \textbf{0.164} & \textbf{6.922} & \textbf{0.149} & \textbf{0.392} & \textbf{2.556} & \textbf{0.104} & \textbf{0.139} & \textbf{7.430} & \textbf{0.050} & \textbf{0.105} \\

\bottomrule
\end{tabular}}
\begin{tablenotes}
\item $\quad\quad$ The best results are in \textbf{bold}, and the second-best are in \underline{underline}.
\vspace{-2mm}
\end{tablenotes}
\end{table*}
\begin{table*}[t]
\centering
\caption{Evaluation Results in \texttt{SNAIL-Radar}}
\label{tab:result_snail}
\resizebox{0.92\textwidth}{!}{
\begin{tabular}{c|ccc|ccc|ccc|ccc|ccc}
\toprule
\multicolumn{1}{l|}{\textbf{}} & \multicolumn{3}{c|}{\texttt{20240113\_1}} & \multicolumn{3}{c|}{\texttt{20240113\_3}} & \multicolumn{3}{c|}{\texttt{20240115\_2}} & \multicolumn{3}{c|}{\texttt{20240123\_2}} & \multicolumn{3}{c}{\texttt{20240123\_3}} \\
\multicolumn{1}{l|}{}          & ATE            & RPE$_t$         & RPE$_r$         & ATE             & RPE$_t$               & RPE$_r$         & ATE             & RPE$_t$         & RPE$_r$         & ATE             & RPE$_t$         & RPE$_r$         & ATE            & RPE$_t$         & RPE$_r$         \\ \midrule

Fast-LIO2 (LiDAR)  & {1.958} & 0.144 & {0.372} & {26.605} & {0.160} & {0.161} & {50.688} & {0.171} & {0.186} & {46.456} & {0.407} & {0.164} & {6.896} & {0.153} & {0.200} \\ \midrule
Go-RIO        & 13.760 & 0.636 & \underline{0.581} & \underline{175.247} & \underline{0.837} & \underline{1.180} & - & - & - & - & - & - & 112.141 & \underline{0.783} & 1.641 \\
PUA-RIO       & \underline{7.964}  & \underline{0.123} & 1.081 & - & - & - & - & - & - & - & - & - & - & - & - \\
EKFRIO-TC     & 12.192 & 0.417 & 1.136 & 278.009 & 2.958 & 1.528 & \underline{346.083} & \underline{3.558} & \underline{1.999} & \underline{511.418} & \underline{1.815} & \underline{0.733} & \underline{41.953} & 1.551 & \underline{0.821} \\

Ours          & \textbf{0.617} & \textbf{0.054} & \textbf{0.180} & \textbf{16.871} & \textbf{0.061} & \textbf{0.099} & \textbf{44.064} & \textbf{0.103} & \textbf{0.092} & \textbf{36.761} & \textbf{0.068} & \textbf{0.094} & \textbf{5.968} & \textbf{0.054} & \textbf{0.105} \\
\bottomrule
\end{tabular}}
\vspace{-5mm}
\end{table*}

\subsection{Scan-to-Submap Registration with Explicit Local Geometry}

Retaining the reliable point in the map enables explicit local geometry assumption in the radar odometry framework. Unlike existing methods that focused solely on rough Gaussian distribution \cite{zhuang20234d, li20234d}, we shift the main attention to the explicit representation of local geometry. For each achieved radar point ${}^\mathrm{R}\mathbf{p}$ at time $t$, five neighbor points are extracted and fitted into the planar model with the local planarity assumption.
The point-to-plane measurement model with its Jacobian is computed with a sample point $\mathbf{p}_{pl}$ in the plane:

\vspace{-4mm}\small
\begin{align}
    h^{pl} &= \mathbf{n}^\top\left(\mathbf{t}(t)+\mathbf{R}(t) \left(\mathbf{R}_\mathrm{IR}{}^\mathrm{R}\mathbf{p}+{}^\mathrm{I}\textbf{t}_\mathrm{IR}\right)-\mathbf{p}_{pl}\right), \\ 
    \mathbf{H}^{pl} &= \left[\mathbf{n}^\top\mathbf{B}(t), \mathbf{n}^\top\frac{\partial \mathbf{R}(t)\left(\mathbf{R}_\mathrm{IR}{}^\mathrm{R}\mathbf{p}+{}^\mathrm{I}\textbf{t}_\mathrm{IR}\right)}{\partial \mathbf{q}(t)}\frac{\partial \mathbf{q}(t)}{\partial \mathbf{x}^q},\mathbf{0},\mathbf{0}\right].
\end{align}
\normalsize
We adaptively reweight these measurement models with the uncertainties, with the uncertainty threshold $\tau_u$: 

\vspace{-4mm}\small
\begin{align}
\label{eq:plane_cov}
    \boldsymbol{\Sigma}_{pl} = \sum_{i=0}^4\left(\cfrac{\tau_u - \text{tr}(\boldsymbol{\Sigma}_{\mathbf{p}_i})}{\sum_{j=0}^4\tau_u -\text{tr}(\boldsymbol{\Sigma}_{\mathbf{p}_j})}\right)^2\boldsymbol{\Sigma}_{\mathbf{p}_i}.
\end{align}
\normalsize
We deem the fitted plane unreliable if $\text{tr}(\boldsymbol{\Sigma}_{pl}) > \tau_{pl}$.
In this case, we switch from the point-to-plane residual to a point-to-distribution residual. The corresponding distribution is estimated with \ac{RCS} augmented fitting as follows:
\begin{equation}
    \mathbf{p}_{pt} = \cfrac{\sum_{i=0}^4 rcs_{\mathbf{p}_i} \cdot \mathbf{p}_i}{\sum_{i=0}^4 rcs_{\mathbf{p}_i}}.
\end{equation}
The point-to-distribution measurement model and its Jacobian are computed using the following equation:

\vspace{-4mm}\small
\begin{align}
    h^{pt} &= w_{rcs}||\mathbf{t}(t)+\mathbf{R}(t) \left(\mathbf{R}_\mathrm{IR}{}^\mathrm{R}\mathbf{p}+{}^\mathrm{I}\textbf{t}_\mathrm{IR}\right)-\mathbf{p}_{pt}||_2, \\ 
    \mathbf{H}^{pt} &= w_{rcs}\frac{\left[\mathbf{B}(t),\frac{\partial \mathbf{R}(t)\left(\mathbf{R}_\mathrm{IR}{}^\mathrm{R}\mathbf{p}+{}^\mathrm{I}\textbf{t}_\mathrm{IR}\right)}{\partial \mathbf{q}(t)}\frac{\partial \mathbf{q}(t)}{\partial \mathbf{x}^q},\mathbf{0},\mathbf{0}\right]}{||\mathbf{t}(t)+\mathbf{R}(t) \left(\mathbf{R}_\mathrm{IR}{}^\mathrm{R}\mathbf{p}+{}^\mathrm{I}\textbf{t}_\mathrm{IR}\right)-\mathbf{p}_{pt}||_2}.
\end{align}
\normalsize
Since RCS encodes material-dependent scattering, we mitigate distributional mismatch by assigning an RCS-informed weight to each correspondence: $w_{rcs}=1/|rcs_\mu - rcs_{{}^\mathrm{R}\mathbf{p}}|$ where $rcs_\mu=\frac{1}{5}\sum_{i=0}^4 rcs_{\mathbf{p}_i}$.
The propagated uncertainties from \eqref{eq:final_unc} are assigned as the measurement covariance of non-planar points.

In addition to spatial measurements, {we incorporate the relative radial velocity information using the following Doppler measurement model and Jacobian}:

\vspace{-4mm}\small
\begin{align}
    h^v &=  \left(\mathbf{R}_\mathrm{IR}{}^\mathrm{R}\mathbf{p}\right)^\top\left[\mathbf{R}(t)^\top\dot{\mathbf{t}}(t)+\boldsymbol{\omega}(t)\times{}^\mathrm{I}\mathbf{t}_\mathrm{IR}\right], \\ 
    \mathbf{H}^v &=   \left(\mathbf{R}_\mathrm{IR}{}^\mathrm{R}\mathbf{p}\right)^\top\left[\mathbf{G}_1,\mathbf{G}_2 ,\mathbf{0},\mathbf{0}\right], \\
    \mathbf{G}_1  &= \mathbf{R}(t)^\top\left(\frac{\partial \dot{\mathbf{t}}(t)}{\partial \mathbf{x}^p}\right), \quad \mathbf{G}_2=\frac{\partial\mathbf{R}(t)\dot{\mathbf{t}}(t)}{\partial \mathbf{q}(t)}\frac{\partial \mathbf{q}(t)}{\partial \mathbf{x}^q}-[{}^\mathrm{I}\mathbf{t}_\mathrm{IR}]^\wedge\frac{\partial\boldsymbol{\omega}(t)}{\partial\mathbf{x}^q}. \nonumber
\end{align}
\normalsize

{Additionally}, we exploit environment-dependent coefficients $w_{pl} = \frac{N_{pl}}{N_{pl}+N_{pt}},w_{pt} = \frac{N_{pt}}{N_{pl}+N_{pt}}$ in each measurement model. $N_{pl}$ and $N_{pt}$ denote the counts of planar and non‑planar points in associated measurements, respectively. These weights regulate the relative influence of planar and non‑planar residuals under varying environmental conditions. In highly structured environments (e.g., urban areas) where planar features are dominant, the weights emphasize explicit planar constraints. Conversely, in unstructured environments where spurious planar detections are more common, the influence of planar measurements is attenuated.


\subsection{IMU Residual with Online Gravity Estimation}
For \ac{IMU} gyroscope readings at time $t$, we leverage the following measurement model and Jacobian:

\vspace{-4mm}\small
\begin{align}
    \mathbf{h}^{gyro} &= \boldsymbol{\omega}(t)+\mathbf{b}_g, \\
    \mathbf{H}^{gyro} &= \left[\mathbf{0}_{3\times 12},\frac{\partial\boldsymbol{\omega}(t)}{\partial\mathbf{x}^q},\mathbf{0}_3,\mathbf{I}_3\right].
\end{align}
\normalsize

For the accelerometer readings, we isolate the gravity direction from each accelerometer sample and constrain orientation by enforcing collinearity between the estimated gravity vector $\mathbf{g}(t)$ and the known world‑frame gravity ${}^\mathrm{W}\mathbf{g}$. Accordingly, the gravity residual is formulated on the unit‑sphere manifold $S^2$:

\vspace{-4mm}\small
\begin{align}
    \mathbf{g}(t) &= \mathbf{R}(t)(\hat{\mathbf{a}}-\mathbf{b}_a) - \ddot{\mathbf{t}}(t), \\
    h^{grav} &= 1-\frac{{{}^\mathrm{W}\mathbf{g}}^\top}{||{}^\mathrm{W}\mathbf{g}||_2}\cdot\frac{\mathbf{g}(t)}{||\mathbf{g}(t)||_2}, \\
    \mathbf{H}^{grav} &= \frac{{{}^\mathrm{W}\mathbf{g}}^\top}{||{}^\mathrm{W}\mathbf{g}||_2} \left(\mathbf{I}-\frac{\mathbf{g}(t)}{||\mathbf{g}(t)||_2}\frac{\mathbf{g}(t)^\top}{||\mathbf{g}(t)||_2}\right) \nonumber \\
    &\quad\cdot\left[\ddot{\mathbf{B}}(t), -\frac{\partial \mathbf{R}(t)(\hat{\mathbf{a}}-\mathbf{b}_a)}{\partial \mathbf{q}(t)} \frac{\partial \mathbf{q}(t)}{\partial\mathbf{x}^q},\mathbf{R}(t),\mathbf{0} \right].
 \end{align}
\normalsize

\subsection{Uncertainty-aware Radar Mapping}

After the state estimation, we transform the newly acquired radar measurements into the world frame and treat them as the mapping candidate. The estimated uncertainties during state estimation are also regarded as a measure of confidence in point selection during the mapping process.
During point insertion in ikd-tree \cite{xu2022fast}, any existing point whose estimated uncertainty exceeds that of the incoming sample is pruned and replaced. As shown in \figref{fig:unc_map}, the iterative replacement of high-uncertainty points with lower-uncertainty observations improves map fidelity, ensuring the preservation of reliable radar measurements.


\section{Experiment}
\label{sec:experiment}



\subsection{Experimental Setup and Baselines}
\label{subsec:exp_setup}

We evaluate our radar-inertial odometry system on three public datasets using 4D Continental radar: \texttt{HeRCULES} \cite{kim2025hercules}, \texttt{SNAIL-Radar} \cite{huai2024snail}, and \texttt{HKUST} \cite{xu2025incorporating}. For \texttt{HeRCULES}, we utilize \texttt{Parking Lot 03}, \texttt{Library 01}, \texttt{Sports Complex 01}, \texttt{Street 01}, \texttt{Bridge 01}, including the campus-scale, city-scale scene with degenerate environment. In \texttt{SNAIL-Radar}, \texttt{20240113\_1}, \texttt{20240113\_3}, \texttt{20240115\_2}, \texttt{20240123\_2}, \texttt{20240123\_3} are exploited for the experiment, covering all sequences in the dataset with SUV. We utilize all three sequences in \texttt{HKUST} for the small-scale indoor evaluation.

We compare ours with recent state-of-the-art radar-inertial odometry algorithms: Go-RIO (scan-to-scan matching) \cite{yang2025ground}, PUA-RIO (point uncertainty-aware scan-to-scan matching) \cite{xu2025incorporating}, and EKFRIO-TC (odometry without scan registration) \cite{kim2025EKFTC}. We also benchmark our algorithm with LiDAR-Inertial Odometry, Fast-LIO2 \cite{xu2022fast}. Unfortunately, due to the absence of LiDAR at \texttt{HKUST}, the comparison with LiDAR-Inertial Odometry was only conducted using \texttt{HeRCULES} and \texttt{SNAIL-Radar}. 
Evaluation is performed in \ac{RMSE} of \ac{ATE} [$\meter$], \ac{RPE} on translation (RPE$_t$) [$\meter\per\meter$] and rotation (RPE$_r$) [$\degree\per\meter$] with evo evaluator \cite{grupp2017evo}.

\subsection{Evaluation Result}
\label{subsec:hercules_exp}
\begin{figure}[t]
    \centering
    \begin{minipage}{0.9\linewidth}
        \centering
        \includegraphics[width=1.0\linewidth, trim=0.3cm 0.1cm 0cm 0.2cm, clip]{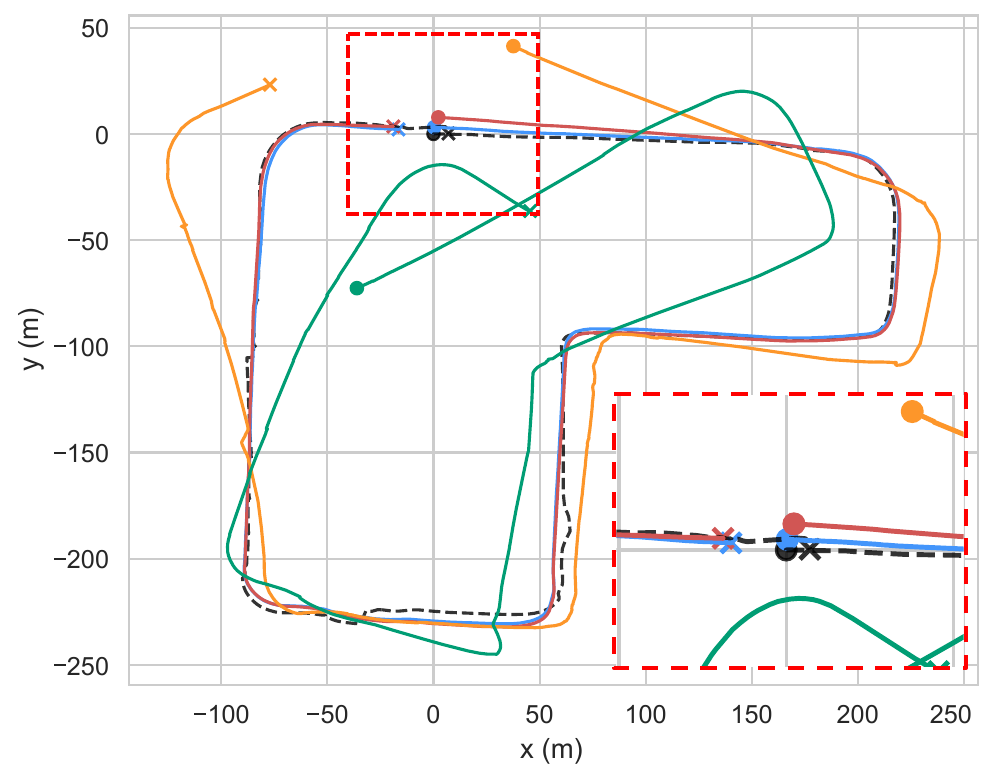}
    \end{minipage}
    \begin{minipage}{1.0\linewidth}
        \centering
        \includegraphics[width=1.0\linewidth, trim=0.0cm 0cm 0.8cm 0cm, clip]{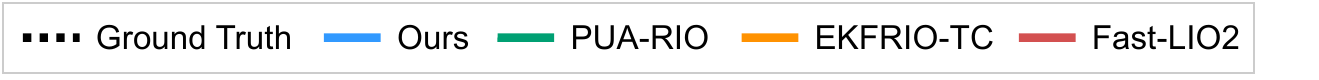}
    \end{minipage}
    \caption{ 
        Estimated trajectory for \texttt{Street}. Our method (blue) shows the best alignment with the ground-truth trajectory (black) and even surpasses LiDAR-Inertial Odometry (red).
    }
    \label{fig:hercules_street}
    \vspace{-6mm}
\end{figure}

\begin{figure}[t]
    \centering
    \begin{minipage}{0.5\linewidth}
        \centering
        \includegraphics[width=1.0\linewidth, trim=0cm 0.5cm 0cm 0cm, clip]{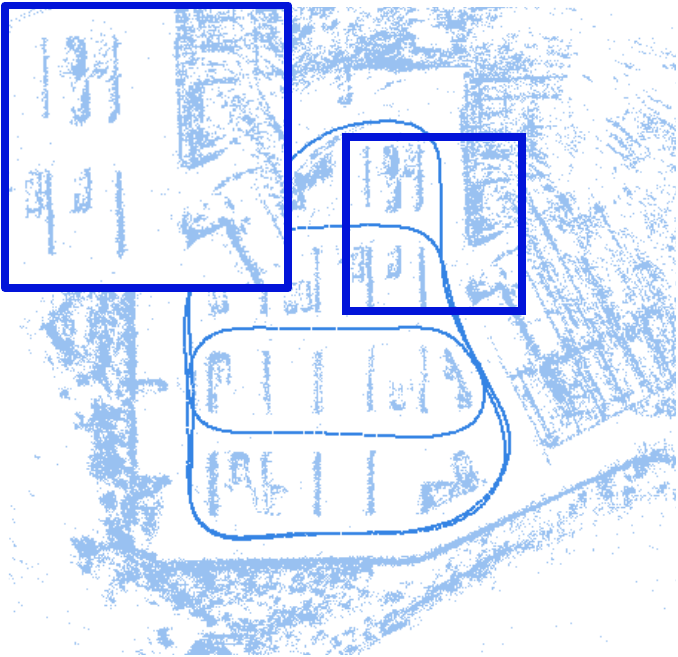}
        \subcaption{Proposed}
    \end{minipage}
    \hfill
    \begin{minipage}{0.48\linewidth}
        \centering
        \includegraphics[width=1.0\linewidth]{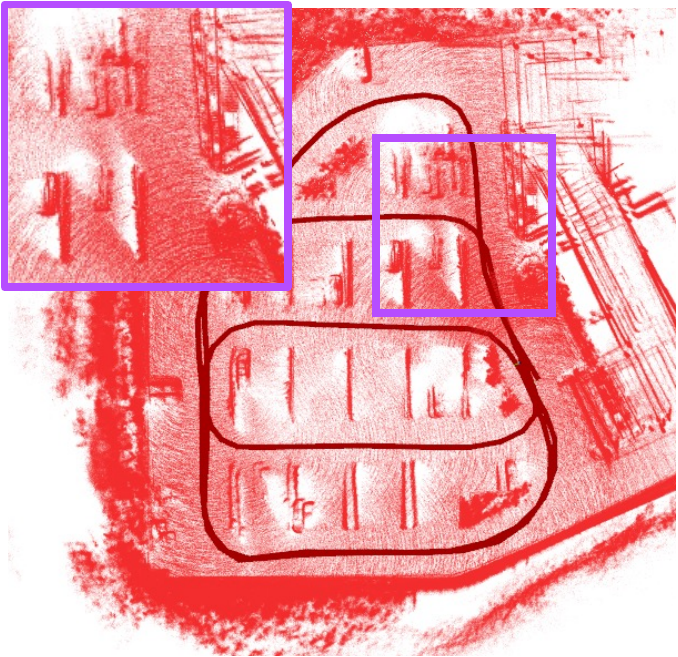}
        \subcaption{Fast-LIO2 (LiDAR)}
    \end{minipage}
    \caption{
        Comparison of odometry and mapping on \texttt{Parking Lot} between the proposed method and LiDAR-Inertial Odometry (Fast-LIO2). Our proposed method yields a well-aligned odometry and mapping result, and also achieves LiDAR-level map consistency.
    }
    \label{fig:parkinglot}
    \vspace{-6mm}
\end{figure}

Quantitative evaluation results in outdoor datasets are summarized in \tabref{tab:result_hercules} and \ref{tab:result_snail}.
Cluster-weighted scan-to-scan matching in Go-RIO suffers from sparse radar point clouds, which leads to incorrect residual weighting. 
Meanwhile, PUA-RIO relies solely on a single source of uncertainty (measurement uncertainty) for weighted scan-to-scan matching. Although prior pose estimate inherently influences landmark accuracy, disregarding pose uncertainty during propagation results in inconsistent landmark generation. Furthermore, dependency on a single-scan landmark sacrifices geometrical consistency, further degrading odometry estimation. As shown in \figref{fig:hercules_street}, this effect is particularly pronounced in \texttt{Street}, where frequent sensor occlusions are caused by traffic congestion.
EKFRIO‑TC's performance remains limited due to the omission of spatial information. This strategy can guarantee accuracy in campus-level sequences, but its shortcomings become pronounced in large-scale urban environments.
Contrastingly, our method achieves superior performance in both ATE and RPE by comprehensively propagating the pose uncertainty and measurement uncertainty and selecting informative points for robust registration.
Moreover, integrating explicit local geometry-guided registration with rigorous uncertainty-aware filtering allows our method to achieve LiDAR-level localization, as illustrated in \figref{fig:parkinglot}.

\begin{table}[t]
\centering
\caption{Evaluation Results in \texttt{HKUST}}
\label{tab:result_hkust}
\resizebox{\columnwidth}{!}{
\begin{tabular}{c|ccc|ccc|ccc}
\toprule
\multicolumn{1}{c|}{} & \multicolumn{3}{c|}{\texttt{Sequence\_1}}                 & \multicolumn{3}{c|}{\texttt{Sequence\_2}}                 & \multicolumn{3}{c}{\texttt{Sequence\_3}}                  \\ 
\multicolumn{1}{c|}{} & ATE            & RPE$_t$         & RPE$_r$         & ATE            & RPE$_t$         & RPE$_r$        & ATE            & RPE$_t$         & RPE$_r$        \\ \midrule

Go-RIO       & 5.771 & 1.348 & \underline{1.346} & 6.146 & 1.808 & 10.955 & 3.901 & 0.903 & \underline{5.464} \\
PUA-RIO      & \underline{0.199} & \underline{0.106} & 1.880 & \textbf{0.246} & \underline{0.167} & \underline{1.839} & \underline{1.064} & \underline{0.202} & 7.542 \\
EKFRIO-TC    & 0.372 & 0.122 & 2.472 & \underline{0.264} & \underline{0.167} & 2.154 & 3.286 & 0.372 & 11.694 \\
Ours         & \textbf{0.146} & \textbf{0.066} & \textbf{0.236} & 0.656 & \textbf{0.064} & \textbf{0.563} & \textbf{0.356} & \textbf{0.044} & \textbf{0.546} \\
\bottomrule
\end{tabular}}
\vspace{-3mm}
\end{table}
\begin{table}[t]
\centering
\caption{Ablation study on uncertainty analysis}
\label{tab:ablation_unc}
\resizebox{1.0\columnwidth}{!}{
\begin{tabular}{c|ccc|ccc|ccc}
\toprule
\multirow{2}{*}{} & \multicolumn{3}{c|}{\texttt{Library 01}} & \multicolumn{3}{c|}{\texttt{20240123\_2}}  & \multicolumn{3}{c}{\texttt{Sequence\_3}} \\
\multicolumn{1}{c|}{} & {ATE}            & {RPE$_t$}         & {RPE$_r$}         & {ATE}            & {RPE$_t$}         & {RPE$_r$}        & {ATE}            & {RPE$_t$}         & {RPE$_r$}        \\ \midrule 
{EKFRIO-TC} & {17.491} & {0.239} & {0.255} & {511.418} & {1.815} & {0.733} & {3.286} & {0.372} & {11.694} \\ 
 w/o Unc. & 21.454 & {0.138} & {0.988} & 72.841 & {0.256} & {0.509}  & 0.697 & {0.213} &{0.597}  \\
 w/o Pose Unc. & {5.114} & {0.139} & {0.976} & 69.458 & {0.231} &  {0.481} & 0.479 & {0.191}  & {0.589} \\
 Full          & \textbf{1.555} & {\textbf{0.090}} & {\textbf{0.164}} & \textbf{36.761} & {\textbf{0.068}} & {\textbf{0.094}} & \textbf{0.356} & {\textbf{0.044}} & {\textbf{0.546}} \\
\bottomrule
\end{tabular}}
\vspace{-3mm}
\end{table}



Unlike other datasets, \texttt{HKUST} is characterized by indoor environments. Our method consistently demonstrates robust performance as represented in \tabref{tab:result_hkust}.
Interestingly, PUA-RIO achieved the second-best performance. Pose uncertainty accumulation is negligible under short-horizon motion, where measurement uncertainty-driven weighting remains more effective.
The absence of geometric constraints reduces EKFRIO-TC’s sensitivity to substantial wall-induced multipath, yielding performance comparable to other baselines.
Go‑RIO incurs the poor \ac{ATE} due to registration without reliability quantification. Additionally, the scarcity of ground returns undermines ground‑optimized filtering.

Notably, ours achieves higher \ac{ATE} on \texttt{Sequence\_2}, which involves high‐dynamic maneuvers of a toy racing car. In this scenario, rapid acceleration and deceleration degrade gravity estimation by the \ac{IMU}, inducing sudden drift in the gravity residual. Despite this, our continuous propagation of both uncertainty effectively mitigates these drifts, resulting in the best performance in \ac{RPE}.




\begin{figure}[t]
    \centering
    \begin{minipage}{\linewidth}
        \centering
        \includegraphics[width=1.0\linewidth]{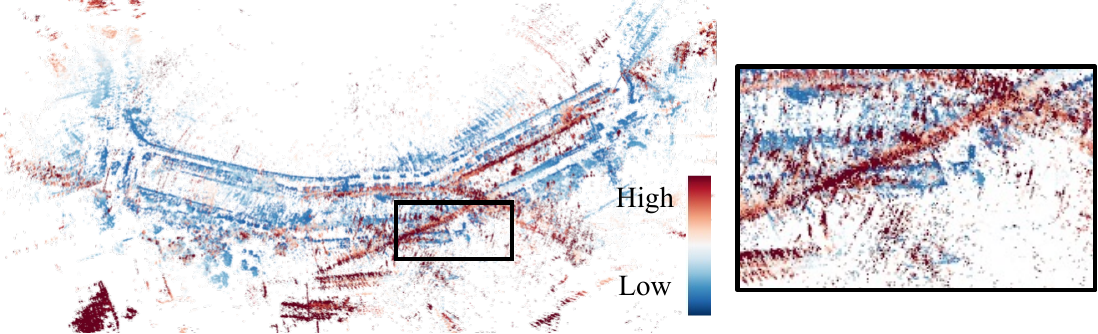}
        \subcaption{Raw mapping without uncertainty}
        \label{fig:before_filter}
    \end{minipage}
    \vfill
    \begin{minipage}{\linewidth}
        \centering
        \includegraphics[width=1.0\linewidth, trim=0.2cm 1.5cm 0 1.5cm, clip]{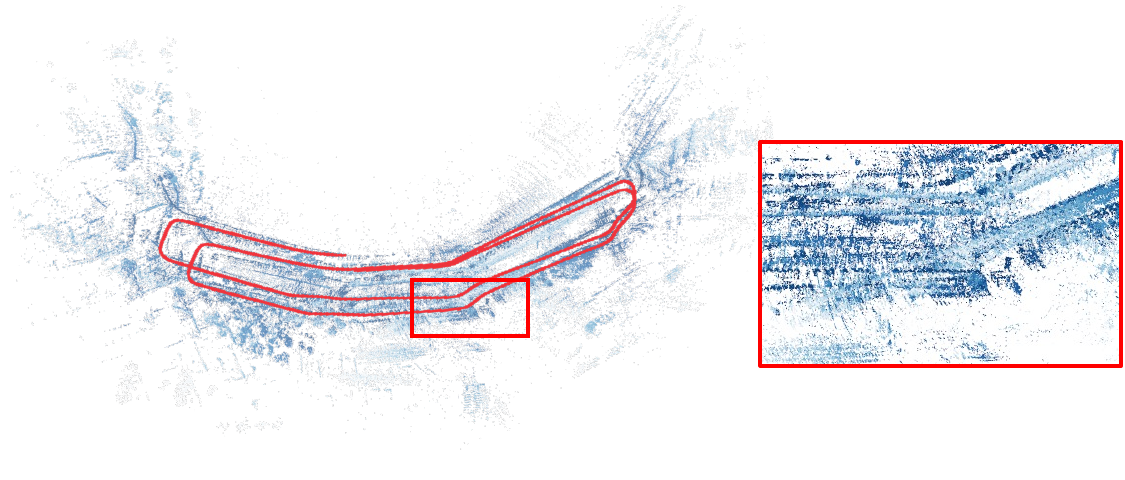}
        \subcaption{Odometry and mapping without pose uncertainty}
        \label{fig:ptunc_filter}
    \end{minipage}
    \vfill
    \begin{minipage}{\linewidth}
        \centering
        \includegraphics[width=1.0\linewidth, trim=0 0.4cm 0 0.7cm, clip]{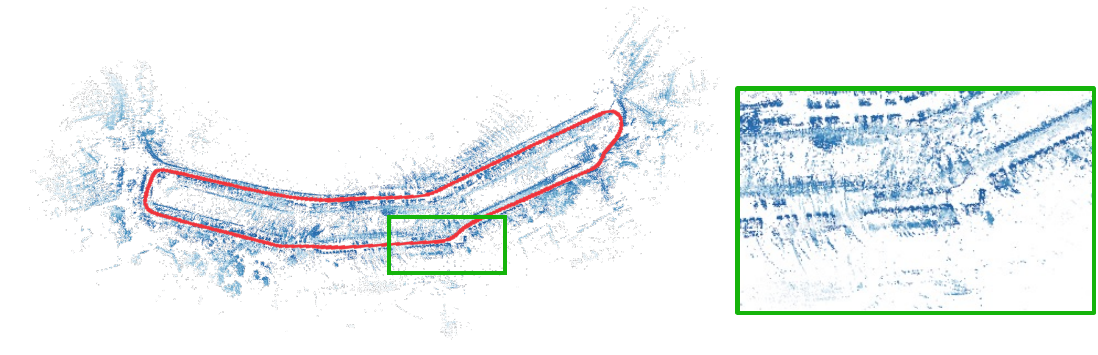}
        \subcaption{Odometry and mapping with proposed full uncertainty}
        \label{fig:after_filter}
    \end{minipage}
    \caption{ 
        (a) Raw mapping result without radar uncertainty in \texttt{Library}. The points are color-coded by their uncertainty, ranging from low (blue) to high (red). Noises from radar returns are projected into the map. (b) Odometry and mapping result without leveraging pose uncertainty propagation. The erroneous measurement still remains on the map. (c) Odometry and mapping result with our full uncertainty module. Our method effectively handles the noise, retaining the reliable points from the measurement.
    }
    \label{fig:unc_map}
\end{figure}

\subsection{Effect of Our Uncertainty Analysis}
\label{subsec:ablation}


To demonstrate the benefit of our complete uncertainty propagation in radar, we compare our full method (Full) against two variants: (i) omitting uncertainty analysis (w/o Unc.), (ii) disregarding pose uncertainty propagation while retaining only the measurement uncertainty (w/o Pose Unc.). As \citet{xu2025incorporating} have already established the effectiveness of measurement uncertainty, our focus is on the impact of jointly integrating point-pose uncertainty. The quantitative results are summarized in \tabref{tab:ablation_unc}. This result clearly shows that our complete uncertainty module outperforms w/o Pose Unc. and uncertainty-ignorant variant w/o Unc., highlighting the necessity of consolidating both pose and measurement uncertainties for a comprehensive analysis in the radar odometry system.
However, in large-scale sequence \texttt{20240123\_2}, the performance gap between w/o Unc. and w/o Pose Unc. is negligible, indicating that without pose-uncertainty propagation, the uncertainty-driven weighting becomes ineffective, and thus yields similar performance to removing uncertainty entirely.
\figref{fig:unc_map} illustrates the qualitative mapping result in \texttt{Library}, and also demonstrates the effectiveness of jointly consolidating both uncertainty for accurate odometry and mapping.

\begin{figure}[t]
    \centering
    \begin{minipage}{\linewidth}
        \centering
        \includegraphics[width=1.0\linewidth, trim=0.0cm 0.4cm 0.2cm 0.0cm, clip]{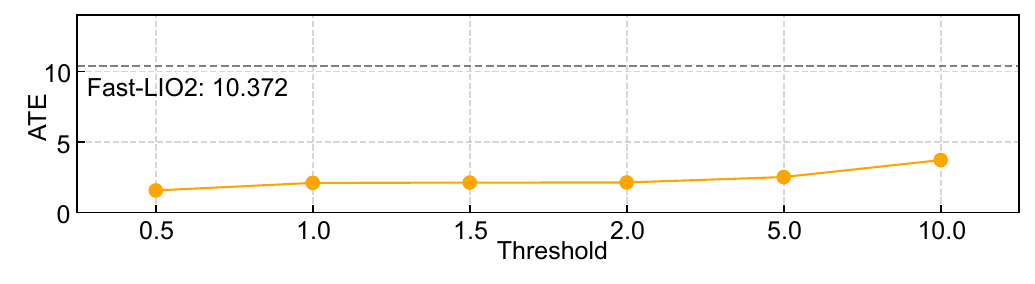}
        \subcaption{\texttt{HeRCULES: Library 01}}
        \label{fig:hercules_ablation}
    \end{minipage}
    \vfill
    \begin{minipage}{\linewidth}
        \centering
        \includegraphics[width=1.0\linewidth, trim=0.0cm 0.1cm 0.2cm 0.1cm, clip]{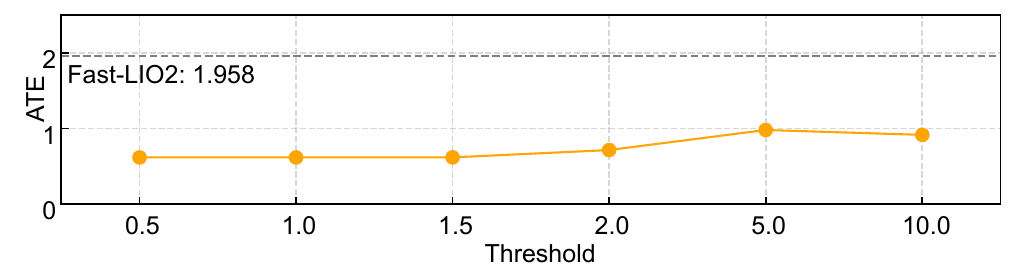}
        \subcaption{\texttt{SNAIL-Radar: 20240113\_1}}
        \label{fig:snail_radar}
    \end{minipage}
    \vfill
    \begin{minipage}{\linewidth}
        \centering
        \includegraphics[width=1.0\linewidth, trim=0 0.2cm 0.2cm 0.0cm, clip]{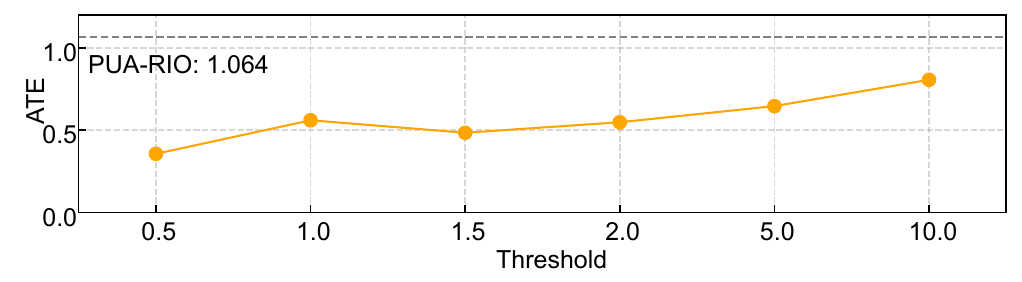}
        \subcaption{\texttt{HKUST: Sequence\_3}}
        \label{fig:hkust_ablation}
    \end{minipage}
    \vspace{-2mm}
    \caption{ 
        Evaluation result with different uncertainty thresholds in the proposed method (orange, ATE). The gray line denotes the result from \underline{second-best} baseline. The proposed method maintains stable accuracy across thresholds and still outperforms other baselines.
    }
    \label{fig:ablation_thresh}
\end{figure}

We also conduct experiments to evaluate odometry accuracy under varying uncertainty thresholds used in the mapping process. (The default setting is 0.5.) As shown in \figref{fig:ablation_thresh}, performance remains stable over a broad range of 
uncertainties and consistently outperforms other baselines across all datasets. This insensitivity to the uncertainty threshold underscores the robustness of the proposed pipeline and reduces the need for heuristic hyperparameter tuning.

\begin{table}[t]
\centering
\caption{Ablation study between uncertainty-aware mapping (Unc. Map) and planar residual (Plane Res.)}
\label{tab:ablation_plane}
\resizebox{1.0\columnwidth}{!}{
\begin{tabular}{c|c||ccc|ccc}
\toprule
\multirow{2}{*}{Unc. Map} & \multirow{2}{*}{Plane Res.} & \multicolumn{3}{c|}{\texttt{Sports Complex}} & \multicolumn{3}{c}{\texttt{20240123\_3}}    \\ 
& & {ATE}            & {RPE$_t$}         & {RPE$_r$}         & {ATE}            & {RPE$_t$}         & {RPE$_r$}       \\ 
\midrule

$\times$ & $\times$      & 9.712  &{0.203}  &   {0.826}      & 17.773 &{0.246}  &   {0.817}       \\
$\times$ & \checkmark    & 20.036 & {0.354} &    {1.249}     & 22.815 & {0.251} & {0.839}        \\
\midrule

\checkmark & $\times$     & {7.531} & {0.192}  &  {0.598}     & {15.707} &{0.247}  & {0.758}    \\
\checkmark & \checkmark   & \textbf{6.922}    &\textbf{{0.149}} & \textbf{{0.392}} & \textbf{5.968} & \textbf{{0.054}} & \textbf{{0.105}}   \\ 
\bottomrule
\end{tabular}
}
\vspace{-5mm}
\end{table}

\subsection{Enabling Planarity Constraints via Uncertainty}

Explicit modeling of local geometry is mainstream in \ac{LiDAR} odometry, but is overlooked in radar-based systems due to the sparsity of returns and high measurement noise. We hypothesize that our uncertainty-guided pipeline can restore the utility of planar constraints within radar-inertial odometry.

\tabref{tab:ablation_plane} indicates that the effectiveness of the planarity residual depends critically on the mapping strategy. We evaluate this ablation on structured environments where planar constraints are prominent. Preserving only low-uncertainty returns and combining them with a geometric residual yields substantial odometry gains, whereas applying a planarity residual directly to an unfiltered map degrades performance because the local planarity assumption is frequently violated. 
Therefore, our central insight is the necessity of the refinement stage during local map construction to enable effective enforcement of planarity constraints.
Exploring other scan-wise refinement strategies that could further improve the validity of local geometric assumptions is left for future work.

\begin{table}[t]
\centering
\caption{Effect of localizability-constrained update}
\label{tab:ablation_loc}
\resizebox{1.0\columnwidth}{!}{
\begin{tabular}{c|ccc|ccc|ccc}
\toprule
\multicolumn{1}{c|}{} & \multicolumn{3}{c|}{\texttt{Library 01}} & \multicolumn{3}{c|}{\texttt{Bridge 01}} & \multicolumn{3}{c}{\texttt{20240123\_2}} \\ 
\multicolumn{1}{c|}{} & {ATE}            & {RPE$_t$}         & {RPE$_r$}         & {ATE}            & {RPE$_t$}         & {RPE$_r$}        & {ATE}            & {RPE$_t$}         & {RPE$_r$}         \\ \midrule
w/o Loc. & 13.082    & {0.140} &   {0.982}    & 45.836    & {0.123} &    {0.643}    & 132.911 & {0.217} & {0.583} \\
Ours    & \textbf{1.555} & \textbf{{0.090}} & \textbf{{0.164}} & \textbf{7.430} & \textbf{{0.050}} & \textbf{{0.105}} & \textbf{36.761}& \textbf{{0.068}} & \textbf{{0.094}} \\ \bottomrule
\end{tabular}}
\vspace{-7mm}
\end{table}

\subsection{Effect of Localizability-constrained IEKF update}
The effect of our localizability-constrained update is presented in \tabref{tab:ablation_loc}. We use sequences with repetitive structures to induce geometric degeneracy. During the experiment, we observed substantial performance degradation in estimation accuracy without constrained update (w/o Loc.), indicating that our method is effective in suppressing drift under these challenging conditions.
The impact of the constrained update becomes evident in \texttt{Bridge} and \texttt{20240123\_2}, which contain repetitive geometrical features on the bridge and tunnel. Although Doppler residual can partially mitigate geometric degeneracy, relying solely on proprioceptive measurements leads to failures in highly dynamic or high-speed scenarios, as shown in the results from EKFRIO-TC in \tabref{tab:result_hercules} and \ref{tab:result_snail}.

\begin{figure}[t]
    \centering
    \includegraphics[width=\columnwidth, trim=0.20cm 0.20cm 0cm 0.2cm, clip]{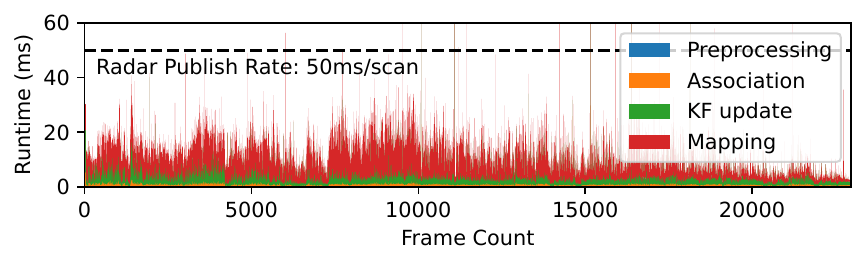}
    \caption{ 
        Runtime analysis in \texttt{20240115\_2}
    }
    \label{fig:runtime}
    \vspace{-6mm}
\end{figure}

\subsection{Computational Time Analysis}
\label{subsec:time}

We analyze the computational cost for \texttt{20240115\_2} in the \texttt{SNAIL-Radar} dataset, which has the longest distance in our experiments. Analysis was conducted on an Intel i7-12700 CPU with 64 GB of RAM. As illustrated in \figref{fig:runtime}, our system can fully support real-time operation (over 20 $\hertz$). These findings indicate that our approach retains sufficient computational room for extension to fusion with other modalities or multi-radar configurations.


\section{Conclusion} 
\label{sec:conclusion} 

In this paper, we presented a robust uncertainty-aware continuous radar-inertial odometry and mapping framework. We proposed a novel continuous uncertainty propagation method that consolidates point and pose uncertainty in the radar system, yielding adaptive covariances that down-weight uninformative observations and preserve a consistent map for robust registration. Building on this, we find out that exploiting the explicit geometric constraints delivers tangible accuracy improvements when combined with our mapping system. We further incorporated a localizability-constrained IEKF to stabilize estimation in challenging scenarios. Extensive experiments on diverse real-world datasets demonstrate consistent improvements over state-of-the-art baselines while maintaining real-time performance.

\bibliographystyle{IEEEtranN}

\bibliography{string-short,references}
\end{document}